\documentclass{svproc}

\usepackage{url}

\usepackage{xcolor}
\usepackage{graphicx}
\usepackage{footmisc}
\usepackage{multicol}
\usepackage{caption}
\usepackage{subcaption}
\usepackage{amssymb}
\usepackage{adjustbox}
\usepackage{booktabs} 
\usepackage{amsmath}
\usepackage[ruled,vlined]{algorithm2e}
\usepackage{hyperref}
\usepackage{graphics} 
\usepackage{multirow}
\usepackage{adjustbox}
\usepackage{tabulary}
\usepackage{multirow, tabularx, longtable, makecell, diagbox}
\usepackage{pifont}

\usepackage{array}

\usepackage{tablefootnote}

\usepackage{multirow}
\newcommand{\ie}{\textit{i.e.}}

\newcommand{\CP}[1]{\ignorespaces}

\begin{document}
\mainmatter              
\title{Multispectral Image Segmentation in Agriculture: A Comprehensive Study on Fusion Approaches}
\titlerunning{MS Image Segmentation in Agriculture: A Comprehensive Study on Fusion Approaches}       
%

\author{Nuno Cunha\and  Tiago Barros \and Mário Reis \and Tiago Marta \and Cristiano Premebida \and Urbano J. Nunes}

\authorrunning{Cunha et al.} 

\tocauthor{Nuno Cunha, Tiago Barros, Mário Reis, Tiago Marta, Cristiano Premebida,  Urbano J. Nunes}

\institute{University of Coimbra, Institute of Systems and Robotics (ISR), Department of Electrical and Computer Engineering (DEEC), Coimbra - Portugal.\\
\email{\tt\small\{nuno.cunha, tiagobarros, mario.reis, tiago.marta,
~cpremebida,~urbano\}@isr.uc.pt}
}

\maketitle
\begin{abstract}

Multispectral imagery is frequently incorporated into agricultural tasks, providing valuable support for applications such as image segmentation, crop monitoring, field robotics, and yield estimation. From an image segmentation perspective, multispectral cameras can provide rich spectral information, helping with noise reduction and feature extraction. As such, this paper concentrates on the use of fusion approaches to enhance the segmentation process in agricultural applications.
More specifically, in this work, we compare different fusion approaches by combining RGB and NDVI as inputs for crop row detection, which can be useful in autonomous robots operating in the field. The inputs are used individually as well as combined at different times of the process (early and late fusion) to perform classical and DL-based semantic segmentation.
In this study, two agriculture-related datasets are subjected to analysis using both deep learning (DL)-based and classical segmentation methodologies. The experiments reveal that classical segmentation methods, utilizing techniques such as edge detection and thresholding, can effectively compete with DL-based algorithms, particularly in tasks requiring precise foreground-background separation. This suggests that traditional methods retain their efficacy in certain specialized applications within the agricultural domain. Moreover, among the fusion strategies examined, late fusion emerges as the most robust approach, demonstrating superiority in adaptability and effectiveness across varying segmentation scenarios. The dataset and code is available at \hyperlink{}{https://github.com/Cybonic/MISAgriculture.git}
\end{abstract}


\section{INTRODUCTION}
In agriculture, autonomous robots are becoming increasingly popular because of the potential benefits they may have on food security, sustainability, resource-use efficiency, reduction of chemical treatments, and optimization of human effort and yield~\cite{9177181}. Alongside this trend, the utilization of multispectral imagery in agricultural applications, including AgRA (Agricultural Robotics and Automation), has become increasingly significant in recent years. Some notable applications of these images include plant disease detection, fruit maturity, and crop production analysis~\cite{jameel2020practical}.

Certain bands, captured at specific frequencies across the electromagnetic spectrum, have the ability to reveal distinct information about plants. Among these bands, the near-infrared (NIR) band holds significance in agricultural tasks (e.g., assessing crop health) as it can effectively highlight chlorophyll absorption and water content in plants. One widely used index that relies on the NIR band is the Normalized Difference Vegetation Index (NDVI), which provides a quantitative measure of vegetation greenness and density. Compared with RGB-only data, incorporating this additional spectral information can enhance the discrimination of different objects and features within images. This enables more accurate identification and classification of crops, improving the process of image segmentation~\cite{yuan2021deep}.

This work focuses on assessing the applicability of fusion approaches using multispectral data for segmentation-related agricultural tasks. Specifically, we investigate two fusion approaches: early fusion and late fusion. Early fusion involves combining the information from multiple sources at the input level before the segmentation process. This means that the data from different sources are merged into a single representation prior to segmentation. On the other hand, late fusion occurs after the segmentation process has been applied to each individual image. The segmentations are obtained independently, and then "fused," or combined, at a later stage. By exploring both early and late fusion techniques, we aim to assess their impact on image segmentation performance and determine which fusion approach yields superior results for the specific objectives of this work.

Through a comprehensive comparative analysis, the aim of this work is to make significant progress in automatic crop-row detection by studying early and late fusion of multispectral data using classical and DL-based segmentation approaches. To accomplish this, this paper brings two key contributions:
\begin{itemize}
\item[$\bullet$] A curated multispectral dataset collected on maize crops using a robotic platform, with crop row annotations;
\item[$\bullet$] An extensive comparison study conducted on both deep learning (DL)-based and classical segmentation methods, focusing on early and late fusion techniques across two distinct datasets. The findings reveal two key insights: First, classical segmentation approaches prove to be competitive with DL-based methods in tasks that involve foreground-background separation, demonstrating their continued relevance in certain applications. Second, late fusion emerges as the most robust fusion approach, showcasing its superior adaptability and effectiveness across various scenarios.
\end{itemize}

\section{RELATED WORK}
\label{sec:related}


Image segmentation is a fundamental task in computer vision, which involves the division of an image into meaningful regions or objects to understand the scene \cite{9356353}\cite{YU201882}\cite{garciagarcia2017review}. In the past, semantic segmentation relied on methods\CP{, which encompassed techniques such as}using thresholding~\cite{SAHOO1988233}, edge-based~\cite{muthukrishnan2011edge} and region-based~\cite{karthick2014survey} \CP{methods}. These methods have the advantage of simplicity and low computational cost.  

On the other hand, convolutional neural networks (CNNs) have revolutionized the field in recent years and are now the most effective technique in pattern recognition application~\cite{lee2015deep}. One of the strongest advantages of using DL in image processing is the reduced need for handcrafted features. These improvements helped agricultural tasks such as disease detection in vines~\cite{KERKECH2020105446}, identification of crops, weeds, and soil~\cite{8460962} through architectures such as encoder-decoder SegNet and Mask R-CNN respectively.

Image Segmentation can \CP{improved the}improve scene understanding however, complex environments require complementary information that multiple modalities can give to better understand the scene \cite{ASVADI201820}. To achieve this goal, fusion methods \CP{must}can be applied which encompass, usually, \CP{, these methods can be divided into} three steps. First, it is necessary to understand which modalities should be fused, then what method should be applied to fuse the information, techniques like addition or average mean, concatenation or ensemble, and finally where should the information be fused along the network~\cite{9000872}\cite{ZHANG2021104042}. Focusing on `where' the information is fused, we highlighted two stages, (i) the early fusion which consists of combining (merging) the data at the input layer \CP{before the modalities enter the network, they are merged}, and (ii) the late fusion which consists of training features separately for each modality and merging them at later layers using methods such as element-wise summation~\cite{valada2017deep}.

\section{MATERIALS AND METHODS}
\label{sec:mm}
This section outlines the methods, tools, and processes employed to conduct the experiments of this work. Firstly, we provide a comprehensive characterization of the study sites and present the technical details of the recorded maize data. Secondly, we formulate the segmentation problem in generic terms and then in a multispectral fusion context by focusing, specifically, on early and late fusion techniques of two distinct information sources.

\begin{figure}[tb]
     \centering
     \begin{subfigure}[b]{1\textwidth}
         \centering
         \includegraphics[width=\textwidth]{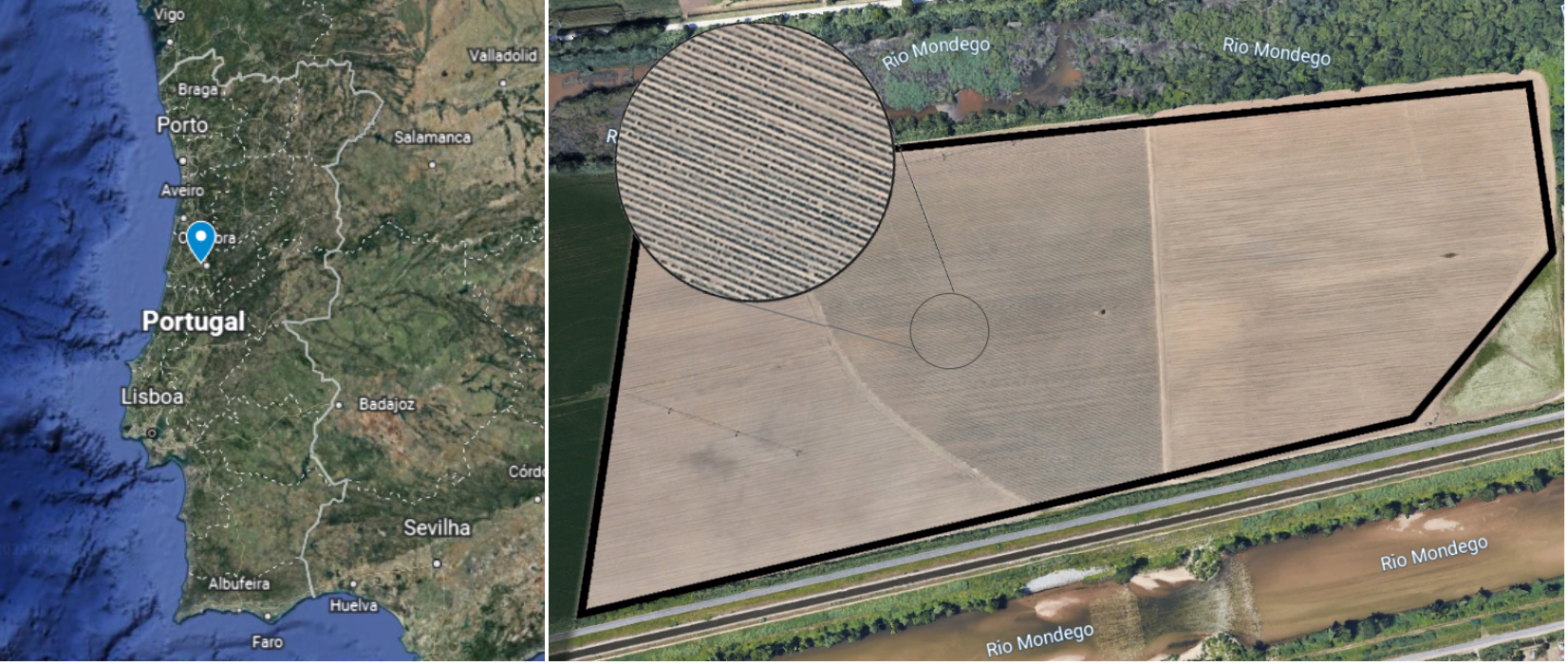}
         \caption{}
         \label{fig:location}
     \end{subfigure}
     ~
     \begin{subfigure}[b]{0.32\textwidth}
         \centering
         \includegraphics[width=\textwidth]{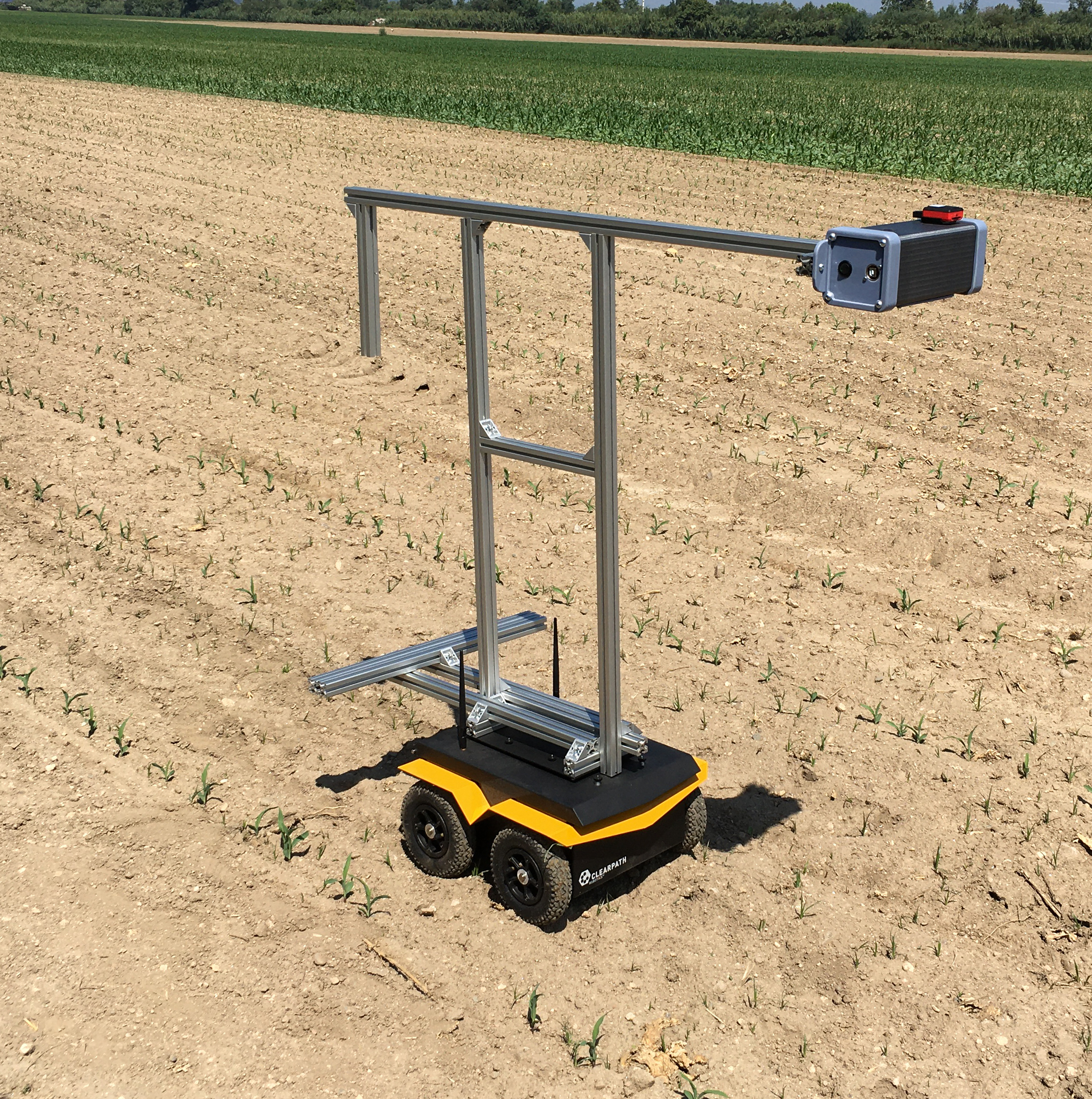}
         \caption{}
         \label{fig:robot}
     \end{subfigure}
     \hfill
     \begin{subfigure}[b]{0.67\textwidth}
         \centering
         \includegraphics[width=\textwidth]{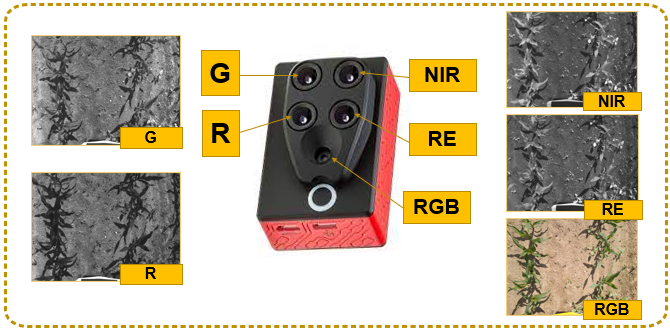}
         \caption{}
         \label{fig:sensor}
     \end{subfigure}
        \caption{Study site and material used to record the dataset, where (a) illustrates the studied maize crop denominated Vargem Grande, (b) is the recording setup with which the dataset was recorded, and (c) is the multispectral sensor with its five sensors.}
        \label{fig:three graphs}
\end{figure}

\subsection{Study Site and Materials}
\label{sec:ssm}

The study was conducted using data collected from a maize crop known as Vargem Grande (VG) located in the Coimbra region, situated in the center of mainland Portugal (see Fig. \ref{fig:location}). The data collection took place during July of 2022, specifically during the early growth stage of the plants. To ensure optimal lighting conditions and minimize shadow interference, the data was collected around midday under sunny weather conditions. 

The multispectral dataset was captured using a Parrot Sequoia multispectral camera\footnote{\label{foot:seq}\href{https://www.parrot.com/assets/s3fs-public/2021-09/sequoia-userguide-en-fr-es-de-it-pt-ar-zn-zh-jp-ko_1.pdf}{Parrot Sequoia User Guide}}. This camera consists of four monochrome sensors (Green, Red, Red Edge, and Near Infrared) along with an RGB sensor (see Fig. \ref{fig:sensor}). To facilitate the data collection process, the camera was mounted on a mobile platform known as the Jackal from Clearpath\footnote{\label{foot:jackal}\href{https://clearpathrobotics.com/jackal-small-unmanned-ground-vehicle/}{Jackal Homepage}}. The camera was positioned 1.2 meters above the ground, with the sensors facing downward (see Fig. \ref{fig:robot}).

To gather the data, the robot was teleoperated in-between the crop rows. Images from all five sensors were captured every two seconds, ensuring a comprehensive dataset for analysis. 

\begin{table}[tb]
\centering
\caption{Specifications of the sensor. Field of
View (FoV)}
{\renewcommand{\arraystretch}{1.15}
\setlength{\tabcolsep}{2.5pt}
\begin{tabular}{l|c|c|c|c|c}
\hline \noalign{\hrule height 1pt}
\multicolumn{1}{l|}{Sensors} &  Band: Center wavelength (width) & Resolution & Focal Length & HFoV & VFoV \\
&[nm] & [px] & [mm] & [º]  & [º] \\ \hline \hline
  Mono- &  G:550(40); R:660(40); RE:735(10); & 1280$\times$960 & 3.98 & 62 & 49 \\
  -chrome & NIR:790(40) & & &\\
  RGB & R,G,B & 4068$\times$3456 & 4.88 & 64 & 50
   \\  \hline 
\noalign{\hrule height 1pt}
\end{tabular}}
\label{tab:cameraspecs}
\end{table}

\subsection{Problem Formulation}

\begin{figure}[tb]
\centering
\includegraphics[width=1\linewidth,keepaspectratio]{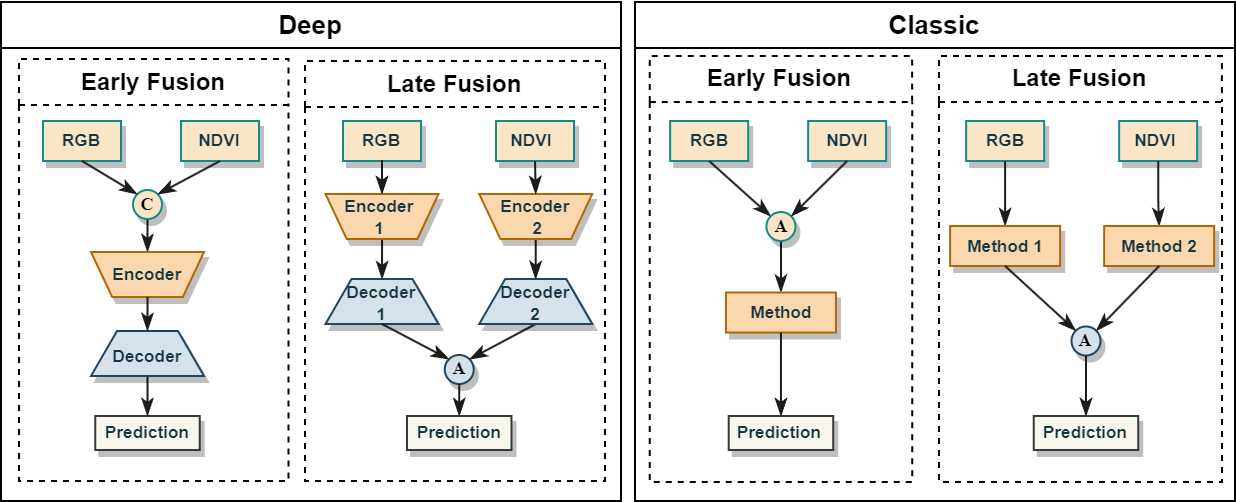}
\caption{Simplified approach of early and late fusion using RGB and NDVI as inputs on deep and classic methods.}\label{fig:fusion}
\end{figure}

Image segmentation involves the task of dividing an image into regions, or objects, based on their shared characteristics. Mathematically,  image segmentation can be defined as a function that maps an input image to a class likelihood mask. 
Thus, let $I$ represent the input image, defined as a three-dimensional array $I = \left[p_{ijk}\right]_{h\times w \times b}$, where $p_{ijk} \in [0,...,255]$ denotes the pixel intensity at coordinates $(i, j, k)$. The image dimensions are given by $h$ (height), $w$ (width), and $b$ (number of spectral bands), with $i \in [1, h]$, $j \in [1, w]$, and $k \in [1, b]$.
To perform image segmentation, we aim to obtain a class likelihood mask $Q$, represented by $Q = [q_{ijk}]_{h\times w \times c}$. Here, $q_{ijk} \in [0,1]$ indicates the likelihood of the pixel at coordinates $(i, j, k)$ belonging to each of the $C = \{1,...,c\}$ segmentation classes, constrained by $\sum_{k=1}^c q_{ijk} = 1$.

In the specific context of this study, we focus on binary segmentation. This means that only one class is considered, resulting in a single-channel likelihood matrix $Q = [q_{ij}]_{h\times w \times 1}$. Hence, the final segmentation mask with a class per pixel $M = [m_{ij}]_{h\times w} \in \{0, 1\}$, is obtained through a threshold-based approach:

\begin{equation}
m_{ij} = 
\begin{cases} 
      1 & \text{if } q_{ij} \geq T \\
      0 & \text{if } q_{ij} < T 
\end{cases}
\end{equation}

\noindent where $T$ is a threshold value chosen to distinguish between the positive and negative classes in the segmentation.

The binary segmentation framework is used to compare classical methods with deep learning (DL)-based approaches using two input modalities: RGB ($I^{RGB}$) and NDVI ($I^{N}$). The RGB image $I^{RGB}$ is defined as a tree-dimensional array $I^{RGB} = [p^{RGB}_{ijk}]_{h \times w \times 3}$, capturing the visible spectrum (400-700\,nm) with the Red, Green, and Blue bands. On the other hand, the NDVI image $I^{N}$ is a two-dimensional array $I^{N} = [p^N_{ij}]_{h\times w}$, representing the Normalized Difference Vegetation Index. The NDVI is calculated as:

\begin{equation}
    I^{N} = \frac{NIR - Red}{NIR + Red}\:,
\end{equation}

\noindent where $Red$ and $NIR$ correspond to specific spectral bands. The Red band lies within the visible spectrum, while the NIR band extends beyond the visible range (700 to 1100\,nm). These bands are particularly valuable for agricultural monitoring, capturing the absorption of chlorophyll in visible light and its reflection in the NIR spectrum.

\subsection{Image Fusion} \label{sec:fusion}

Fusion, in the context of image segmentation, refers to the integration of information derived from diverse sources into a unified representation. The fusion process can be applied at various stages, depending on the segmentation methods employed~\cite{valada2016towards}. In this study, we specifically investigate two fusion approaches: early fusion and late fusion.

\subsubsection{Early Fusion}
In the context of image processing, early fusion involves the merging of information at the input level, specifically within the pixel space. In this study, early fusion is employed using two different approaches: classical segmentation methods and DL-based segmentation methods.

In the comparison between classical and DL-based methods, the representation of early fusion varies depending on the approach used. Specifically, when employing classical approaches, the RGB image $I^{RGB}$ is transformed into a grayscale representation denoted as $I^{Gr} = [p^{Gr}_{ij}]_{h\times w}$. This conversion is achieved using the standard formula:

\begin{equation}
    p^{Gr}_{ij} = 0.299\,p^R_{ij} + 0.587\,p^G_{ij} + 0.114\,p^B_{ij}\;,
\end{equation}
\noindent where $p^R_{ij}$, $p^G_{ij}$, and $p^B_{ij}$ represent the pixel intensities of the Red, Green, and Blue bands at the coordinate $(i,j)$, respectively, with $i \in [1,h]$ and $j \in [1,w]$. The resulting grayscale image $I^{Gr}$ \CP{is of dimensions}has dimensions given by $h \times w$.

For classical approaches, the fused representation $I^{Ec}$ is obtained by computing the pixel-wise mean between the NDVI image $I^{N}$ and the grayscale image $I^{Gr}$:

\begin{equation}
    p^{Ec}_{ij} = \frac{p^{N}_{ij} + p^{Gr}_{ij}}{2}\;,
\end{equation}

\noindent here, $p^{N}_{ij}$ and $p^{Gr}_{ij}$ represent the pixel intensities of the NDVI and grayscale images at the $(i,j)$ coordinate, respectively. The resulting fused representation $I^{Ec}$ is an image of dimensions $h \times w$. On the other hand, when employing DL-based segmentation methods, the fused representation $I^{Ed}$ is obtained by channel-wise concatenation of the RGB image $I^{RGB}$ and the NDVI image $I^{N}$. This is represented as:

\begin{equation}
    I^{Ed} = [I^{RGB}, I^{N}] = \left[ p^{RGB}_{ijk} \, | \, p^{N}_{ijk} \right]_{h \times w \times 4}\;
\end{equation}
\noindent where $p^{RGB}_{ijk}$ and $p^{N}_{ijk}$ represent the pixel intensities of the RGB and NDVI images at the $(i,j,k)$ coordinate, respectively. The resulting fused representation $I^{Ed}$ is a tensor \CP{of}with dimensions $h \times w \times 4$, where the first three channels correspond to the RGB image and the fourth channel corresponds to the NDVI image.

\subsubsection{Late Fusion}
Early fusion involves merging information at the input space, while late fusion performs the merging at the output space. In this study, late fusion is achieved by computing the pixel-wise weighted sum of the class likelihoods of each model before the final class decision. 
 


In the context of a late fusion framework, the segmentation process involves two input images: $I^N$ and $I^{RGB}$. Each image is individually processed through a segmentation model, generating respective output likelihood masks: $Q^N = [q^{N}_{ij}]_{h \times w \times 1}$ and $Q^{RGB} = [q^{RGB}_{ij}]_{h \times w \times 1}$, where, $q^{N}_{ij}$ and $q^{RGB}_{ij} \in [0,1]$ represent the likelihood of the positive class at the pixel coordinates $(i,j)$.

The fused representation is obtained by computing a pixel-wise weighted sum of the likelihoods from both segmentation models. Hence, the fused likelihood $q^L_{ij}$ at the pixel coordinates $(i,j)$ is calculated using the following formula:

\begin{equation}
q^L_{ij} = \alpha \cdot q^N_{ij} + \beta \cdot q^{RGB}_{ij}\;,
\end{equation}
\noindent where $\alpha$ and $\beta$ are weights that can be adjusted to balance the contribution of each likelihood according to the models' performance. By controlling the values of $\alpha$ and $\beta$, the fusion process can be fine-tuned to achieve optimal segmentation results based on the strengths of the individual models.

\section{EXPERIMENTAL EVALUATION}  \label{sec:experiments}

The evaluation section in this study provides a comprehensive assessment of early and late fusion techniques within a multispectral image segmentation framework applied to the AgRA domain. The section outlines the datasets used for evaluation, describes the implementation details and evaluation metrics employed, and presents a thorough discussion of the quantitative and qualitative results obtained.

\subsection{Datasets}

The proposed approaches undergo evaluation using \CP{two datasets:}primarily the maize crop dataset (referred to as VG) described in Section \ref{sec:ssm}. Complementary, \CP{ and additional data} a dataset collected from vineyards are used to assess cross-domain generalization capability. For the VG dataset, a total of 532 images were recorded for each of the five sensors (R, G, RE, NIR, and RGB). The images were aligned and cropped to a final size of $1100 \times 825$, and for evaluation purposes, they were resized to $240 \times 240$. The dataset was then split into an 80/20 ratio for training and testing, respectively. Regarding the vineyard data, the dataset encompasses images of $240 \times 240$ from three distinct vineyards. The evaluation follows the approach proposed in \cite{BARROS2022106782}, employing a cross-validation method that involves training on data from two vineyards and testing on the third. \CP{All} Relevant information about the datasets can be found in Table~\ref{tab:datasetinfo}.

\begin{table}[tb]
\centering
\caption{Dataset information, where B,G,R,RE and NIR represent Blue, Green, Red, Red-edge and Near-infrared, respectively.}
\label{tab:datasetinfo}
{\renewcommand{\arraystretch}{1.2}
\setlength{\tabcolsep}{5.5pt}
\begin{tabularx}{\textwidth}{l|c|ccc}
\hline \noalign{\hrule height 1pt}
\multicolumn{1}{c|}{Dataset} & Vargem Grande      & \multicolumn{1}{c|}{Qta Baixo}     & \multicolumn{1}{c|}{ESAC}     & Valdoeiro   \\ \hline \hline
Sample Size (Train/Test) & 532 (425/107)               & 150           & 189                 & 120   \\
Bands  & R, G, RE, NIR, RGB & \multicolumn{3}{c}{B, G, R, RE, NIR, Thermal}  \\  
Dimensions Fusion & 1100$\times$825         & \multicolumn{3}{c}{240$\times$240}   \\  \hline 
\noalign{\hrule height 1pt}
\end{tabularx}}
\end{table}

\subsection{Implementation Details}

This section outlines the implementation details of both the classical and DL-based segmentation approaches. 

\subsubsection{Classical Approach}
Three classical segmentation methods were employed: Otsu's thresholding\footnote{\label{foot1}\href{https://docs.opencv.org/4.x/d7/d4d/tutorial_py_thresholding.html}{OpenCV Image Thresholding - Otsu's thresholding.}}, edge-based\footnote{\label{foot}{\href{https://scikit-image.org/docs/stable/auto_examples/applications/plot_coins_segmentation.html}{Edge-based and Region-based segmentation - Canny edge-detector and  Watershed transform.}}}, and region-based$^{\ref{foot}}$ techniques. For Otsu's thresholding, the opencv \emph{threshold} function with an automatic threshold value was utilized to perform the segmentation. In the case of edge-based segmentation, the Canny edge detector was employed to detect the edges of the objects, with further processing to fill the contours and remove small objects from the segmented image. Lastly, a region-based segmentation was performed by generating an elevation map using the Sobel gradient, determining markers for background and plants based on gray value histograms, and then applying the watershed transform to fill regions of the elevation map with those markers.

\subsubsection{Deep Learning Approaches}
In this study, two distinct DL-based segmentation models were utilized: SegNet\footnote{\href{https://github.com/trypag/pytorch-unet-segnet}{SegNet GitHub Implementation}} and DeepLabV3\footnote{\href{https://pytorch.org/hub/pytorch_vision_deeplabv3_resnet101/}{DeepLabV3 Pytorch}}. SegNet employs an encoder-decoder architecture, where the input is gradually encoded to a latent space and then gradually decoded to an output mask. In contrast, DeepLabV3 upsamples the latent representation in fewer steps.

Both models were implemented using the PyTorch~\cite{paszke2019pytorch} framework and executed on a hardware setup consisting of an NVIDIA GEFORCE GTX 3090 GPU and an AMD Ryzen 9 5900X CPU with 64 GB of RAM. The training process utilized the AdamW optimizer~\cite{loshchilov2017decoupled} with a learning rate of 1e-3 for VG and approximately 1e-4 and 1e-5 for vine models. The Binary Cross-Entropy with Logits Loss (BCEWithLogitsLoss) function was employed to calculate the loss, and the outputs (logits) were passed through a sigmoid activation function to obtain the final probabilities.

\subsection{Evaluation Metrics}

The performance of the segmentation methods was evaluated using several metrics, including pixel accuracy (acc), $F_{1}$ score, and Intersection over Union (IoU). These metrics provide insights into the accuracy and quality of the segmentation results. The pixel accuracy is defined as:
\begin{equation}\label{eq:acc}
\text{acc} = \frac{{TP + TN}}{{TP + FP + TN + FN}},
\end{equation}
\noindent where TP, TN, FP, and FN represent True Positives, True Negatives, False Positives, and False Negatives, respectively. The $F_{1}$ score is calculated as:
\begin{equation}\label{eq:f1}
F_{1}\text{ score} = \frac{{2 \times TP}}{{2 \times TP + FP + FN}}.
\end{equation}

\noindent The $IoU$ is computed as :
\begin{equation}\label{eq:iou}
    IoU = \frac{{\text{{Area of Intersection}}}}{{\text{{Area of Union}}}}\;,
\end{equation}
where \emph{Area of Intersection} refers to the number of overlapping pixels between the predicted mask and ground truth mask: $A \cap B = \{p_{ij} : p_{ij} \in A \text{ and } p_{ij} \in B\}$, where $p_{ij}$ denotes a pixel at coordinate (i, j), while $A$ and $B$ represent the ground truth mask and the predicted mask, respectively.  The \emph{Area of Union} represents the total number of pixels encompassed by both prediction and ground truth masks, including the overlapping region: $A \cup B = \{ p_{ij} : p_{ij} \in A \text{ or } p_{ij} \in B \}$, where $p_{ij}$ denotes a pixel at coordinate (i, j), while $A$ and $B$ represent the ground truth mask and the predicted mask, respectively.

\begin{table}[pt]
\centering
\caption{Segmentation performance on the Maize (VG) and Vine (Qta. Baixo, ESAC, and Valdoeiro) datasets, employing classical approaches such as Otsu Threshold (OST), Edge-based, and Region-based, as well as DL-based approaches including SegNet and DeeplabV3. Each method is evaluated with four scores: RGB and NDVI individually, and both modalities fused using early and late fusion techniques. The performance scores are presented in percentage [\%], with the \textbf{best score} highlighted in bold and the \underline{second-best} scores underlined.}
{\renewcommand{\arraystretch}{1.15}
\setlength{\tabcolsep}{1.2pt}
\begin{tabularx}{\dimexpr\textwidth}{cc|ccc|ccc|ccc|ccc|ccc}
\hline \noalign{\hrule height 1pt}
& & \multicolumn{3}{c|}{Maize} & \multicolumn{9}{c|}{Vine} \\ \cline{3-14} 
\multicolumn{2}{l|}{\diagbox{\tiny{Method}}{\tiny{Dataset}}}
&\multicolumn{3}{c|}{VG} &
\multicolumn{3}{c|}{Qta. Baixo}  & 
\multicolumn{3}{c|}{ESAC} & 
\multicolumn{3}{c|}{Valdoeiro} &  \multicolumn{3}{c}{Average} \\ 
& & Acc. & F1 & IoU & Acc. & F1 & IoU & Acc. & F1 & IoU & Acc. & F1 & IoU 
& Acc. & F1 & IoU\\   \hline \noalign{\hrule height 1pt}
\multirow{4}{*}{\rotatebox[origin=l]{90}{OTS}} & 
RGB & 74.4 & 28.3 & 16.7 
& 57.6 & 41.9 & 27.6 
& 79.1 & 52.7 & 40.2 & 
74.5 & 38.6 & 26.0 &
71.4 &  40.4 & 27.6\\
& NDVI &76.1 & 32.3 & 19.6 
& 84.1&  61.7&  46.1        
& 65.6 & 49.7 & 34.8       
& 92.4 & 67.8 & 56.0 &
\underline{79.6} &  \underline{52.9}&  \underline{39.1}\\  \cline{2-17}  
&  Early F. & 75.6 & 34.1& 20.9
& 71.6 &52.7 &37.7
& 71.4 & 55.5 & 41.2
& 89.5 & 65.3 & 52.7&
77.1 & 51.9 &   38.1\\ 
& Late F.& 67.5 & 33.7 & 20.8 
& 83.0& 67.5& 44.7 
& 89.5&  66.8 & 42.9
& 91.6 & 81.6 &56.7 &
\textbf{82.7} & \textbf{62.4} & \textbf{41.3} \\  \hline \hline
\multirow{4}{*}{\rotatebox[origin=c]{90}{Edge-b.}} & RGB & 
76.6 & 12.9 & 6.9 
& 69.8 & 15.5 & 8.5 
& 78.1 & 26.3 & 15.4 
& 86.9 & 22.7 & 13.1 &
\underline{77.9} & 19.4 & 10.5\\
& NDVI & 75.6 &13.0 &7.0
& 70.3& 16.8 &9.3
& 61.1& 18.7& 10.4
&  94.2& 48.6 &33.7&
75.3 & \underline{24.3} & \textbf{15.1} \\  \cline{2-17}
& Early F.&  84.1& 21.0& 11.9
& 76.8 &16.3& 8.9
& 65.4 &14.8 &8.0
& 93.9 &39.3 &25.8&
\textbf{80.1}& 22.9& 13.7 \\ 
& Late F.& 70.1 & 14.0 & 7.6
& 64.8 & 18.9 & 10.6
& 71.1& 25.5 & 14.8
& 87.6& 39.1& 25.1&
73.5 &\textbf{24.4}& \underline{14.5}\\  \hline \hline
\multirow{4}{*}{\rotatebox[origin=c]{90}{Region-b.}} & RGB & 
 84.1 & 21.0& 11.9
& 78.3& 47.4& 33.1
& 78.9& 44.3& 33.2
& 85.3& 44.6& 31.2&
81.7& 39.3& 27.4\\
& NDVI & 82.2& 12.4& 6.7
& 89.0& 67.9& 52.5
& 76.0& 50.9& 37.3
& 97.2& 76.3& 63.8
& \underline{86.1} & 51.9 & \underline{40.1}\\ \cline{2-17}
&  Early F. & 76.7& 19.0& 10.5
& 81.3 & 52.7& 37.0
& 87.6 & 63.5& 49.8
& 97.3 & 77.6&  65.2 
& 85.8 & \underline{53.2}& \textbf{40.6}\\ 
& Late F. & 81.8& 25.3& 14.7
& 93.1& 69.3& 34.9
& 92.1& 48.9& 24.6 
& 98.6& 82.7& 45.2
& \textbf{91.4} &\textbf{56.6} &29.9\\   \hline \noalign{\hrule height 1pt} 
\multirow{4}{*}{\rotatebox[origin=c]{90}{SegNet}} & 
RGB & 96.2 & 87.1 & 78.5 &
84.9 & 52.1 & 35.6 & 
73.1 & 41.9 & 27.7  & 
92.1 & 58.0 & 41.3 &
86.6 & 59.8 & 45.8 \\
& NDVI & 95.6 & 85.3 & 76.1 & 
85.9 & 64.4 & 48.4 &
78.5 & 51.4 & 36.8 &  
93.9  & 67.1 & 51.7 & 
\textbf{88.5} &\textbf{67.1} & \textbf{53.3} \\  \cline{2-17}
& Early F. & 96.8 & 89.3 & 80.8 & 
81.1 & 42.1 & 26.7 & 
81.4 & 50.5 & 33.9 & 
94.3 & 61.0 & 43.9 &
\underline{88.4} & 60.7 & 46.3 \\ 
& Late F. & 96.1 & 86.9 & 78.6 & 
86.2 & 56.4 & 40.4 & 
75.9 &  46.9 & 33.0  &
93.4 & 61.4 & 45.7 &
87.9 & \underline{62.9} & \underline{49.4} \\  \hline \hline
\multirow{4}{*}{\rotatebox[origin=c]{90}{DeeplabV3}} & 
RGB & 96.5 & 87.9 & 79.8 &
81.8 & 35.6 & 21.8 &
82.0 & 45.0 & 30.1 &
91.0 & 56.2 & 39.7 &
\underline{87.9} & \underline{56.1} & \underline{42.9}\\
& NDVI & 95.9 & 86.0 & 77.3 &
87.3 & 59.2 & 42.2 &
77.7 & 33.8 & 21.4 &
89.1 & 44.2 & 28.8 &
87.5 & 55.8 & 42.4\\  \cline{2-17}
&  Early F. & 97.3 & 89.2 & 81.2 & 
82.1 & 31.0 & 18.7 &
79.9 & 37.3 & 23.8 &
89.9 & 52.8 & 36.4 &
87.3 & 52.6 & 40.0\\
& Late F. & 96.6 & 87.7 & 80.2 &  
85.3 & 47.5 & 32.0 &
81.0 & 39.0 & 25.9 &
92.5 & 58.0 & 42.3 &
\textbf{88.9} & \textbf{58.1} & \textbf{45.1} \\  \hline
\noalign{\hrule height 1pt}
\end{tabularx}
\label{tab:classical}}
\end{table}


\subsection{Results and Discussion}

This section presents the experimental results for both classical and DL-based segmentation methods, comprising both quantitative and qualitative assessments.
The qualitative results are organized in Table~\ref{tab:classical}, while the visual representations of the segmentation masks are illustrated in Fig~\ref{fig:masks}. Each segmentation approach is evaluated in four distinct methods: first, with the RGB and NDVI modalities individually, followed by the modalities fused using early and late fusion techniques, as described in Section \ref{sec:fusion}.  The results were obtained with the segmentation threshold $T = 0.5$, for the late fusion results, both models were given an equal contribution: \ie\; $\alpha = 0.5$ and $\beta = 0.5$.  

\subsubsection{Classical vs DL-based}
\begin{figure}[tb]
\centering
\includegraphics[width=1\linewidth,keepaspectratio]{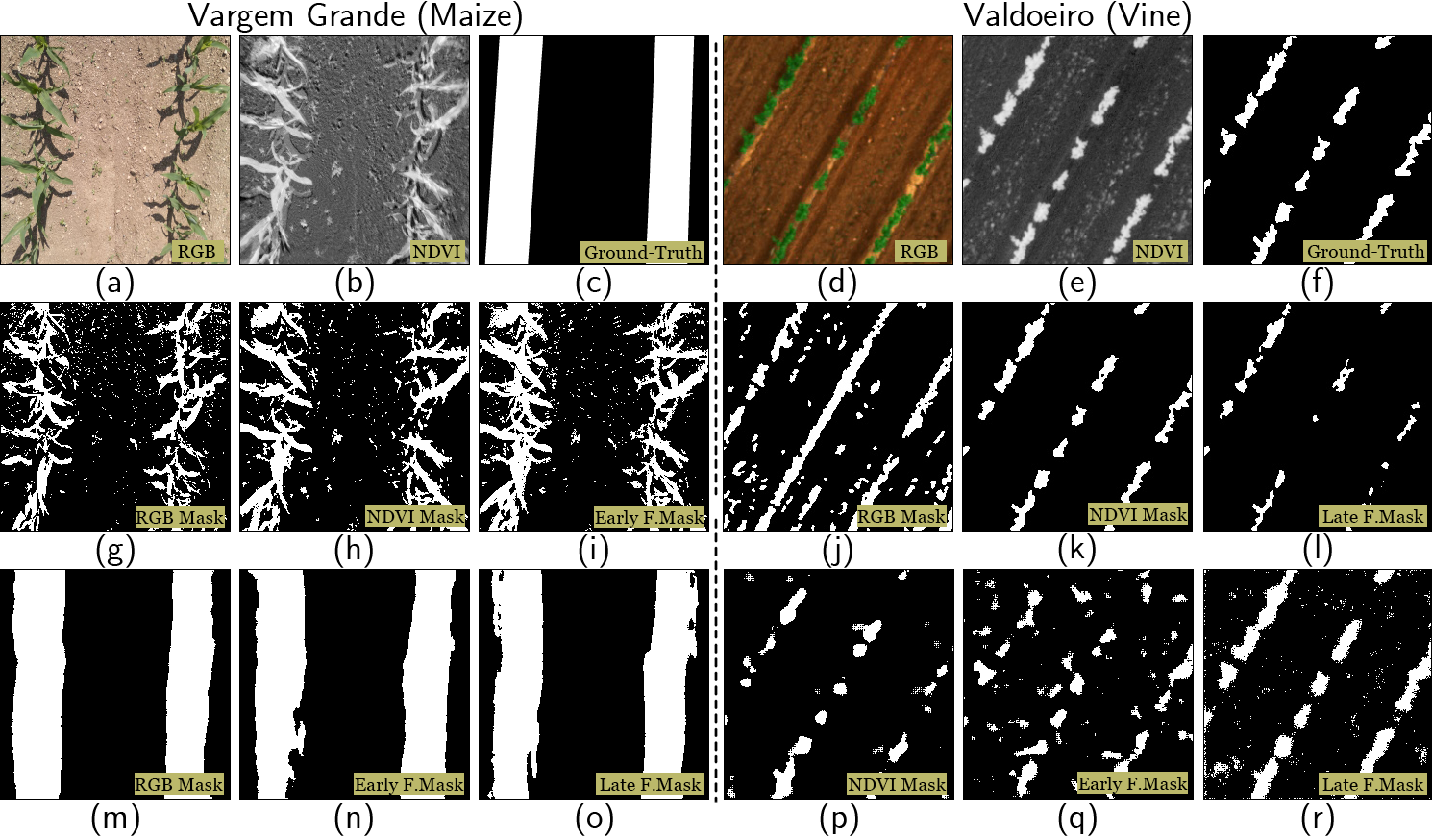}
\caption{Qualitative segmentation results of both VG and vineyard dataset. The images (a) to (f) (top row), represent respectively the RGB, NDVI and ground-truth masks. Images (g) to (l) (middle row) represent segmentation masks generated by classical approaches. And finally, images (m) to (r) (bottom row) represent segmentation masks generated by SegNet. More specifically, images (g) to (i) were generated by Otsu, while images (j) to (l) were generated with a region-based method.}
\label{fig:masks}
\end{figure}

In this work, we employ classical unsupervised and supervised DL-based segmentation methods. The classical methods demonstrate to perform well on tasks where the primary  objective is to separate foreground from background, as is the case of the Vineyard dataset, where the goal is to segment individual plans. In such case, unsupervised approaches are competitive with DL-based approaches, offering the advantage of simplicity and lower complexity. However, in segmentation tasks that involve identifying spatial regions, containing both foreground and background, such as the Maize dataset, where the objective is to detect the plant rows, supervised DL-based approaches show a clear advantage due to their ability  to learn spatial information. The results obtained in our experiments consistently confirm this, as depicted in Table~\ref{tab:classical} and Fig~\ref{fig:masks}.

\subsubsection{Fusion vs No-Fusion}
The results consistently show that late fusion either achieves the best performance or ranks a close second, distinctly outperforming early fusion. This superiority means that, on average, extracting features from individual modalities first and then fusing them at a later stage yields better results compared to one model from both modalities combined.

Upon analyzing the average results, it becomes evident that late fusion capitalizes on the model with the highest performance. By averaging the outputs of both models, late fusion is able to reduce the noise associated with the lesser-performing model. However, this method also has a downside: valuable information from the best-performing model may be diluted or lost. Thus, while late fusion leverages the strengths of both models to enhance overall robustness, finding the right balance in the contributions of each model becomes crucial.  One potential approach to achieve this balance is to weight the contributions based on their respective performance. Investigating this weighted fusion strategy offers an interesting avenue for future work.

\subsubsection{Runtime Analysis}
In terms of computational performance, DL methods demand a considerable amount of time to execute due to the intensive computations involved. In our case, the maximum runtime reached approximately twenty-five minutes for the entire training process, specifically during late fusion, where the batch size supported by the hardware was limited to 32 (VG) and 16 (Vine). In contrast, classical methods demonstrate the opposite behavior, being significantly faster and achieving results within a minute.

\section{CONCLUSIONS}
\label{sec:conclusions}

This work studies the impact of fusion (combining) approaches of multispectral data in segmentation tasks \CP{in agriculture}applied domains related to digital-precision agriculture and agricultural robotics. The study was conducted on both classical and DL-based segmentation methods, \CP{.  To achieve this objective,}where the experimental part is supported by two datasets \CP{were utilized}: a dataset of vineyards and a dataset of maize crops, recorded and curated specifically for this study. 

The experimental findings show  two principal observations: First, classical segmentation methods, utilizing techniques like thresholding and edge detection, are competitive against DL-based approaches in tasks requiring foreground-background separation. This highlights their continued applicability in specialized scenarios. Second, late fusion, where individual modalities are processed and then fused, emerges as the most robust approach, demonstrating its superior adaptability across various experimental conditions. These insights offer valuable guidance for both current applications and future research in segmentation algorithms.

\section*{ACKNOWLEDGMENTS}
This work has been supported by the project GreenBotics (ref. PTDC/EEI-ROB/2459/2021), founded by Fundação para a Ciência e a Tecnologia (FCT), Portugal. It was also partially supported by FCT through grant UIDP/00048/2020 and under the PhD grant with reference 2021.06492.BD.


\bibliographystyle{spmpsci}
\bibliography{ref.bib}

\end{document}